\documentclass{article}

\usepackage[english]{babel}

\usepackage[letterpaper,top=2cm,bottom=2cm,left=2.5cm,right=2.5cm,marginparwidth=1.75cm]{geometry}

\usepackage{graphicx}
\usepackage[colorlinks=true, allcolors=blue]{hyperref}
\usepackage{microtype}
\usepackage{libertine}
\usepackage[libertine]{newtxmath}
\usepackage[scaled=0.94]{zi4}
\usepackage{authblk}
\usepackage{natbib}
\DeclareMathOperator*{\argmax}{arg\,max}
\DeclareMathOperator*{\argmin}{arg\,min}
\usepackage[linesnumbered, ruled, vlined]{algorithm2e}
\usepackage{subcaption}
\usepackage{multirow}
\usepackage[labelfont=bf,textfont=bf]{caption}

\usepackage{tikz}
\usetikzlibrary{shapes,positioning}

\newcommand*\cir[1]{\tikz[baseline=(cir.base)]{
    \node[shape=circle,draw,fill=gray!15,inner sep=2pt] (cir) {#1};}}

\title{Rank-based Non-dominated Sorting}
\author[1,2,3]{Bogdan Burlacu}
\affil[1]{University of Applied Sciences Upper Austia}
\affil[2]{Heuristic and Evolutionary Algorithms Laboratory}
\affil[3]{Josef Ressel Centre for Symbolic Regression}

\begin{document}
\maketitle

\begin{abstract}
Non-dominated sorting is a computational bottleneck in Pareto-based multi-objective evolutionary algorithms (MOEAs) due to the runtime-intensive comparison operations involved in establishing dominance relationships between solution candidates. In this paper we introduce Rank Sort, a non-dominated sorting approach exploiting sorting stability and ordinal information to avoid expensive dominance comparisons in the rank assignment phase. Two algorithmic variants are proposed: the first one, \emph{RankOrdinal} (RO), uses ordinal rank comparisons in order to determine dominance and requires $O(N)$ space; the second one, \emph{RankIntersect} (RS), uses set intersections and bit-level parallelism and requires $O(N^2)$ space.
We demonstrate the efficiency of the proposed methods in comparison with other state of the art algorithms in empirical simulations using the NSGA2 algorithm as well as synthetic benchmarks. The \emph{RankIntersect} algorithm is able to significantly outperform the current state of the art offering up to 30\% speed-up for many objectives.
C\texttt{++} implementations are provided for all algorithms.
\end{abstract}

\section{Introduction}
Multiobjective optimization problems (MOP) consist of finding the best
compromise or trade-off between a collection of competing objectives. Formally,
a MOP can be defined as: \begin{equation} \text{Minimize}~F(X) \text{ subject
    to } x \in X \end{equation} where $F : X \to Y$, $X \in \mathbb{R}^n$ is
    the \emph{decision variable space} (also called a \emph{feasible set}) and
    $Y \in \mathbb{R}^m$ is the \emph{objective function value space}.

Pareto optimality is an established formalism used to approach this type of
problem, based on the concept of dominance.
Given two vectors $y, z \in \mathbb{R}^m$, it defines the following dominance
relationships:
\begin{align}
    \text{equality} & & y = z \Leftrightarrow y_i = z_i, \forall i=1,...,m \\
    \text{strong dominance} & & y \prec z \Leftrightarrow y_i < z_i, \forall i=1,...,m \\
    \text{weak dominance} & & y \preceq z \Leftrightarrow y_i \leq z_i, \forall i=1,...,m~\text{and}~y \neq z
\end{align}

Thus, if $y \preceq z$ then we say that \emph{$y$ dominates $z$} (\emph{$z$ is
dominated by $y$}). If $y \npreceq z$ and $z \npreceq y$ then $y$ and $z$ are
\emph{mutually non-dominated}. A \emph{Pareto front} represents a set of
mutually non-dominated solutions.

Note that $\prec$ is a \emph{strict partial order} and $\preceq$ is a
\emph{partial order} on the objective space $R^m$. This allows to more
generally describe Pareto relationships in terms of ``predecessors'' and
``successors'' as will be needed later on when Rank Sort is introduced.

When optimizing multiple objectives, it is often the case that a solution
cannot be further optimized with respect to any objective without worsening at
least one of the others. The concept of \emph{Pareto optimality} represents an
intuitive interpretation of this situation. If $x^*$ is a \emph{Pareto optimal}
solution, then there is no other solution $x \in X$ such that $x \preceq x^*$.

Evolutionary algorithms (EA) are one of the most popular methods for solving
multi-objective optimization problems (MOP). In Pareto-based MOEAs,
non-dominated sorting (NDS) refers to the process of dividing the population
into Pareto fronts which are subsequently used to guide parent selection.

In this paper we present an approach to non-dominated sorting which can
substantially reduce the runtime costs of the parent selection step. This is
motivated by the fact that, due to its transactional nature and the need to
carry rank information from one individual to the next, non-dominated sorting
does not lend itself well to parallelization and can substantially increase the
runtime of MOEAs.

The remainder of the paper is organized as follows:
Section~\ref{sec:related-work} gives an overview of existing algorithms from
the literature. Section~\ref{sec:ranksort} discusses the proposed approach,
Section~\ref{sec:results} provides experimental results and
Section~\ref{sec:conclusion} is dedicated to conclusions.

\section{Related work}\label{sec:related-work}

Improvements in the asymptotic complexity of non-dominated sorting algorithms
are mainly driven by new techniques to reduce (or eliminate altogether) the
number of Pareto dominance calculations necessary to divide the population into
Pareto fronts.

We distinguish between two main categories of approaches: those that perform
direct dominance comparisons and those that perform sorting in order to determine
the Pareto ranking of the population.

\subsection{Inference-based approaches}

Inference-based methods avoid unnecessaneuripsry dominance comparisons by exploiting
the transitive property of dominance relationships (e.g., if $a \prec b$ and $b
\prec c$, then $a \prec c$) in order to improve the asymptotic complexity and
runtime performance.

\smallskip\noindent\emph{\sf \emph{Deductive sort (DS)}}~\citep{McClymont2012}
iterates over the solutions in natural order and maintains a strict order of
assessment and comparison to ensure that all dominated solutions are discarded
and not incorrectly added to the wrong front.
Once a front has been filled, all solutions assigned to that front are ignored
and the process is repeated until all solutions are assigned to a front.

\smallskip\noindent\emph{\sf \emph{Efficient non-dominated
sorting}}~\citep{Zhang2015} avoids unnecessary dominance comparisons by
comparing a solution that needs to be assigned to a front only with the other
solutions that have already been assigned. Two algorithmic variants are
proposed, differing in the search strategy used to find if any solution in a
given front dominates the current solution. Two search strategies are proposed:
sequential search (ENS-SS) and binary search (ENS-BS).

\smallskip\noindent{\sf \emph{Hierarchical Non-dominated Sorting
(HS)}}~\citep{Bao2017} sorts all solutions lexicographically and then uses
successive rounds of comparison between the first solution and the succeeding
solutions to establish dominance relationships. Dominated solutions are
discarded in the current round of comparison so they will not be compared again
with the first solution or other solutions which are non-dominated with the
first solution. Once a solution is ranked it is removed from the comparison and
added to its respective front.

\subsection{Sort-based approaches}

In contrast with inference-based approaches where dominance is explicitly
considered, sort-based approaches rely on ordinal ranking information (in some
form or another) in order to determine dominance relationships between solutions.
This typically involves a more expensive preprocessing step where the solutions
are sorted w.r.t. every objective.

The most notable examples of sort-based approaches are Best Order Sort
(BOS)~\citep{Roy2016} and Merge Non-dominated Sort (MNDS)~\citep{Moreno2020}.
Both approaches share the same principneuripsle of using a solution's \emph{dominance
set} (i.e., the set of solutions dominating the current solution) to update its
rank. The dominance set is obtained as the intersection between the sets of
solutions ranked better than the current solution according to each objective.
The ranking is obtained by stable sorting -- a key characteristic which is also
employed by our algorithm.

A key difference between BOS and MNDS is given by the actual implementation of
the set intersection procedure for obtaining the dominance set. BOS uses a
linked list to store a solution's dominance set while MNSE uses a bitset. Both
algorithms are deeply integrated with merge sort which they use in the sorting
step and employ various other auxiliary data structures to keep track of
already ranked solutions or duplicates (MNDS).

\smallskip\noindent\emph{\sf \emph{Best Order Sort
(BOS)}}~\citep{Roy2016}\footnote{Implementation available at
\url{https://github.com/proteekroy/Best-Order-Sort}} sorts the population
according to each objective preserving lexicographical order. The resulting
ranking is used to compute the dominance set for each solution. Finally, a
solution's rank becomes the rank of the worst-ranked solution in the dominance
set plus one.

\smallskip\noindent\emph{\sf \emph{Merge Non-dominated Sort
(MNDS)}}~\citep{Moreno2020}\footnote{Available in the open-source framework
jMetal (\url{https://github.com/jMetal/jMetal})}. Employs a stable-sorting
algorithm (merge sort) to rank solutions according to each objective and uses
the ranking to construct a solution's dominance set as the set of solutions
that dominate the current solution in every objective. Then, the rank of the
current solution is updated according to the worst rank in the dominance set.
Prior to rank assignment, the algorithm removes duplicate solutions from the
population and keeps track of them in a separate list. Finally, the ranks of
the solutions are obtained based on the dominance sets and duplicates are
inserted again with their corresponding rank.

A summary of the runtime complexity and space requirements of the described algorithms is given in Table~\ref{tab:nds-complexity}.

\begin{table}
    \caption{Summary of algorithm time and space complexity. For DS, the worst
    case is $O(MN^3)$ cf. Mishra and Buzdalov~\cite{Mishra2020}.}\label{tab:nds-complexity}
    \begin{tabular}{lllll}
        \textbf{Algorithm} & \textbf{Best} & \textbf{Worst} & \textbf{Space}\\
        \hline
        DS     & $O(MN^2)$      & $O(MN^3)$ & $O(N)$\\
        HS     & $O(MN\sqrt N)$ & $O(MN^2)$ & $O(N)$\\
        MNDS   & $O(MN\log N)$  & $O(MN^2)$ & $O(N^2)$\\
        ENS-SS & $O(MN\sqrt N)$ & $O(MN^2)$ & $O(1)$ \\
        ENS-BS & $O(MN\log N)$  & $O(MN^2)$ & $O(1)$ \\
        RO     & $O(MN\log N)$  & $O(MN^2)$ & $O(N)$\\
        RS     & $O(MN\log N)$  & $O(MN^2)$ & $O(N^2)$\\
    \end{tabular}
\end{table}

\section{Rank-based Non-dominated Sorting}\label{sec:ranksort}

Similar to MNDS and BOS, the Rank Sort algorithm uses the concept of a
``dominance set'' to compute solution ranks. However, while MNDS and BOS define
the dominance set as the set of predecessors of the current solution, Rank Sort
defines the dominance set as the set of \emph{successors} of the current
solution. Here, the notions of ``predecessor'' and ``successor`` correspond to
the $\preceq$ and $\succeq$ Pareto relationships, respectively.

Considering the dominance set as the set of successors brings an important
advantage: during rank update, it is sufficient to consider \emph{equally
ranked successors} of the current solution (considering provisional ranks
iteratively computed by the algorithm). Consequently, the rank update becomes a
simple \emph{increment operation} of the successor's rank. This insight
simplifies the rank update phase and reduces the number of operations such that
the total number of rank updates will be equal to the sum of ranks in the final
Pareto front assignment.

Due to this rank update rule, the algorithm's behavior can be interpreted as
establishing each new Pareto front by moving dominated individuals from the
current Pareto front into the new one.

We present two variants that differ in their space complexity:
\begin{itemize}
    \item \emph{RankOrdinal} (RO) requires $O(N)$ space by avoiding set
        intersections. Instead, it takes the smallest set of successors w.r.t.
        any objective $k=1,...,M$ as the dominance set:
        \begin{equation}
            D_\text{ordinal}(S_i) = D_k(S_i), k = \argmin_l |D_l(S_i)|\label{eq:dom-ordinal}
        \end{equation}

    \item \emph{RankIntersect} (RS) requires $O(N^2)$ space for storage of
        dominance sets and employs bit-level parallelism to speed up the
        computation of set intersections. The dominance set is defined as:
        \begin{equation}
            D_\text{intersect}(S_i) = \bigcap D_k(S_i), k=1,...,M\label{eq:dom-intersect}
        \end{equation}
\end{itemize}

Here, $D_k(S_i)$ represents the set of successors w.r.t. objective $k$:
\begin{equation}neurips
    D_k(S_i) = \{ S_j~|~\text{rank}(S_i) = \text{rank}(S_j)\text{ and }S_j \succeq S_i\}\label{eq:k-dominance-set}
\end{equation}

Node that Equation~\ref{eq:k-dominance-set} should be taken in the context of
iterative rank assignment within the algorithm's inner loop, where the ranks
are not yet final.

\emph{RankIntersect} is the faster algorithm overall however its space
requirements might make it less suitable for resource-constrained environments,
where \emph{RankOrdinal} might be preferable.

\subsection{RankOrdinal Algorithm}\label{subsec:rank-ordinal}

\emph{RankOrdinal} (RO) uses the comparison of ordinal ranks to establish
dominance. Let $i\in 1,...,N$ be an index over the solutions $S_i$, let $k \in
1,...,M$ be an index over the objectives, and let $F_k(i)$ be a function
returning the value of the $k$-th objective for solution $S_i$.

For every objective $k$, we compute the permutation vector $\mathbf{p}_k$ and
ordinal rank vector $\mathbf{r}_k$:
\begin{align}
    \mathbf{p}_k &= \big[p_k(1)~\ldots~p_k(N)\big]^T~\text{\small such that}~F_k(i) \leq F_k(i+1)\label{eq:permutation}\\
    \mathbf{r}_k &= \big[r_k(1)~\ldots~r_k(N)\big]^T~\text{such that}~r_k\left(p_k(i)\right) = i\label{eq:rank}
\end{align}

Let $\mathbf{P} = [\mathbf{p}_k^T] \in \mathbb{N}^{N\times M}$ and $\mathbf{R}
= [\mathbf{r}_k^T] \in \mathbb{N}^{N\times M}$ be the permutation matrix and
ordinal rank matrix, where $k=1,...,M$.

It is clear from Eq. \ref{eq:permutation} that for every $\mathbf{p}_k \in
\mathbf{P}$, the leftmost solution $S_{p_k(1)}$ is always non-dominated and in
general, $S_{p_k(i)} \nsucceq S_{p_k(j)}$ for every $i < j$.
Conversely, a solution $S_{p_k(i)}$ can only dominate other solutions that come
after it in the permutation.
Therefore, it will be efficient to examine solutions left-to-right in the
partial permutation where $S_{p_k(i)}$ is farthest from the left, such that
there remain fewer successors to be examined.
%

By virtue of Eqs. \ref{eq:permutation}, \ref{eq:rank}, the matrix $\mathbf{R}$
helps infer a dominance relationship between two solutions:
\begin{align}
    \text{$S_i$ non-dominated with $S_j$},~\text{if}~r_k(i) > r_k(j),\exists k=1,...,M\label{eq:non-dominance-infer}\\
    S_i \preceq S_j,~\text{if}~r_k(i) < r_k(j), \forall k=1,...,M\label{eq:dominance-infer}
\end{align}

Non-dominated sorting algorithms assign solutions to their corresponding Pareto
fronts based on the \emph{domination rank} value (see for
example~\citep{Deb2002}). Testing for dominance in permutation order gives our
approach the advantage of a natural rank update mechanism.
\begin{equation}
    S_i \preceq S_j\text{ and } \text{rank}(i) = \text{rank}(j) \implies \text{rank}(j) \gets \text{rank}(i) + 1
    \label{eq:rank-update}
\end{equation}
When the rank of $S_j$ is updated, the rank of $S_i$ will have already been
updated at a previous iteration. The entire algorithm is described in
pseudocode in Algorithm~\ref{alg:rankord}. The outer loop iterates over
solution indices given by $\mathbf{p}_1$ (the first column of $\mathbf{P}$).
The inner loop of the algorithm will iterate over the smallest set of
successors, as defined by~\ref{eq:dom-ordinal}. The corresponding column of
$\mathbf{P}$ and the start index for the partial permutation are identified at
line 6 in Algorithm~\ref{alg:rankord}.
The rank assignment procedure is performed at lines 9--11.
The comparison of ranks $r_k$ is actually an element-wise comparison between
the columns of matrix $\mathbf{R}$ corresponding to the two solutions (fast in
practice due to vectorization and the fact that the elements are integers).

\subsubsection{Solution equality}\label{subsubsec:solution-equality}

Note that due to its reliance on sorting, ordinal ranking is unable to detect
the situation when two solutions are equal in all objectives. To handle this
case it is necessary to extend for example Eqs.~\ref{eq:dominance-infer}
and~\ref{eq:rank-update} to explicitly check for equality. This can be cheaply
implemented by e.g. hashing objective values in the preprocessing phase.
However, it is a better design choice to separate duplicates handling logic
from non-dominated sorting.
\begin{algorithm}[ht]
    \small
        \SetKwFunction{rankOrd}{RankOrdinal}
    \SetKwFunction{stableSort}{StableSort}
    \KwIn{$S_1,...,S_N$}
    \KwOut{Pareto fronts $F_1, F_2, ... $}
    \SetKwProg{Fn}{Function}{:}{}
    \Fn{\rankOrd{$S_1,...,S_N$}}{
        $\mathbf{P} \gets$~\stableSort{$S_1,...,S_N$}\; 
        $\mathbf{R} \gets$~calculate using~Eq.~\ref{eq:rank}\;
        \textbf{rank} $\gets$ zero-initialized array of size $N$\;
        \ForEach{$S_i \in \mathbf{p}_1$}{
            $k \gets \argmax_l r_l(i)$\;
            \If{$r_k(i) = N$}{
                \textbf{continue}\tcp*{$S_i$ cannot dominate any other}
            }
            \ForEach{$S_j \in \mathbf{p}_k$}{
                \If{\upshape{$\text{rank}(i) = \text{rank}(j)$}\upshape{\textbf{ and }}$S_i \preceq S_j$ cf. Eq~\ref{eq:dominance-infer}}{
                        $\text{rank}(j) \gets \text{rank}(j) + 1$\;
                    }
                }
            }
        $\mathbf{F} \gets$ list of fronts of size $\max_i\text{rank}(i$)+1\;
        \ForEach{\upshape{$i$ from $1$ to $N$}}{
            append $i$ to front $F_{\text{rank}(i)}$\;
        }
        \textbf{return} $\mathbf{F}$\;
    }

    \caption{\emph{RankOrdinal}}\label{alg:rankord}
\end{algorithm}

\subsubsection{Correctness of the algorithm}\label{subsubsec:rankord-correctness}

The permutations of solutions w.r.t. every objective are generated by calling a
sorting procedure. It is important to note that the algorithm is correct only
when the sorting procedure is stable, as otherwise Eqs.
\ref{eq:non-dominance-infer}, \ref{eq:dominance-infer} would not be reliable.
This issue is apparent in certain extended definitions of dominance such as
$\varepsilon$-dominance~\citep{Laumanns2002} where in the absence of stability
there can be no guarantee about the order in which $\varepsilon$-dominated
solutions are added to their respective fronts.


Many non-dominated sorting algorithms explicitly make use of sorting in order
to avoid the cost of dominance comparisons in at least one
dimension~\citep{Zhang2015,Roy2016,Bao2017,Zhou2017,Xue2020,Moreno2020}.
However, only a few explicitly consider stability~\citep{Roy2016,Moreno2020}.
While this is almost never a problem in empirical runs, it can be a potential
source of inconsistency in MOEAs, for example if two equal solution candidates
are placed on the Pareto front(s) in different relative order due to sorting
instability.
Therefore, in some particular cases, use of an unstable stable sorting
procedure may introduce an undesirable dependency to an implementation detail
that can cause unexplained behavior or an inability to reproduce results (i.e.,
runs with a fixed seed). This was also observed by~\citep{Buzdalov2018} who
called this phenomenon ``bug-compatible'' non-dominated sorting.

\subsubsection{Computational complexity}

Since the preprocessing part has a fixed cost (one lexicographic sorting step
followed by $M-1$ regular sorting steps, its complexity is $O(M N \log N$).
For the rank assignment part of the algorithm, we first consider the total
number of iterations performed. Regardless how little work is done per
iteration (e.g., when the conditions allow to skip over an already ranked
solution), this quantity is important in determining the final complexity of
the algorithm.

As shown in Algorithm~\ref{alg:rankord}, the rank assignment procedure consists
of two nested loops: the outer loop iterates over all the solutions $S_i$ in
the order given by $\mathbf{p}_1$, while the inner loop iterates over the
successors of $S_i$ in the permutation $\mathbf{p}_k$ where $k=\argmax_l r_l(i)$.

\subsubsection*{Worst case}
The worst case occurs when the input is degenerate, namely when the points to be sorted either belong to a single front or belong to individual fronts.
\begin{itemize}
    \item When the solutions are non-dominated and belong to a single Pareto front, this corresponds to the unique situation where matrix $\mathbf{P}$ has the form:
        \begin{equation}
            \mathbf{P} = \begin{bmatrix}
                1 & 2 & ... & N\\
                N & N-1 & ... & 1
            \end{bmatrix}^T
        \end{equation}
        In this case, the size of dominance set $D_\text{ordinal}(S_i)$ will be:
        \begin{equation}
            |D_\text{ordinal}(S_i)| = \min(i, N-i)
        \end{equation}
        Therefore, the total number of iterations will be:
        \begin{equation}
            \text{total iterations} = 2 \cdot \sum_{i=1}^{N/2} i = 2 \cdot \frac{\frac{N}{2} \left( \frac{N}{2}+1 \right)}{2} \approx \frac{N^2}{4}
        \end{equation}

    \item When each solution $S_i$ belongs to its own unique front, this corresponds to the unique situation where matrix $\mathbf{P}$ has the form:
    \begin{equation}
        \mathbf{P} = \begin{bmatrix}
            1 & 2 & ... & N\\
            1 & 2 & ... & N
            \end{bmatrix}^T
    \end{equation}

    In this case, the size of dominance set $D_\text{ordinal}(S_i)$ will be:
    \begin{equation}
        |D_\text{ordinal}(S_i)| = N-i
    \end{equation}

    The total number of iterations in the worst case will be the sum:
    \begin{equation}
        \text{total iterations} = \sum_{i=1}^N i = \frac{N(N+1)}{2} \approx \frac{N^2}{2}
    \end{equation}
\end{itemize}
Note that the results above hold for $M > 2$ since adding more columns to $\mathbf{P}$ would not influence the resulting fronts because of the degenerate configuration of the first two columns. Therefore, \emph{RankOrdinal} has the worst-case complexity of $O(M N^2)$.

\subsubsection*{Best case} In the following we make the argument that the best-case complexity of \emph{RankOrdinal} is $O(M N\log N)$. We consider the problem of finding the Pareto set of randomly selected points in the unit hypercube.
\begin{itemize}
    \item A full dominance check is always precluded by the condition $\text{rank(i)} = \text{rank}(j)$ in Algorithm~\ref{alg:rankord} line 10. Since the algorithm works by incrementing the ranks of dominated solutions within the current Pareto front (in order to ``move'' them to a new front), this condition will be true a finite number of times, and this number is proportional to the size of the Pareto fronts.
    \item Pareto front size is bounded in expectation by generalized harmonic numbers and thus scales in proportion to the natural logarithm
~\citep{Yukish2004}, \citep{Koeppen2005}. Therefore, the number of time the condition $\text{rank}(i) = \text{rank}(j)$ is true will also be proportional to the logarithm of $N$, bounding the dominance checks.
\end{itemize}

    Additionally, it can be shown that the expected minimum of $M$ uniform i.i.d. random variables is
        \begin{equation}
            \mathbb{E}\left[ \min(X_i) \right] = \frac{1}{M+1}
        \end{equation}
    For random objective values sampled from the unit hypercube, the total number of iterations (regardless of the amount of work done inside the inner loop and without considering the rank equality condition) will be $\frac{N^2}{M+1}$.

\begin{figure*}[ht]
    \caption{RankOrdinal total number of iterations and total number of rank comparisons}\label{fig:rankord-complexity}
    \includegraphics[width=\textwidth]{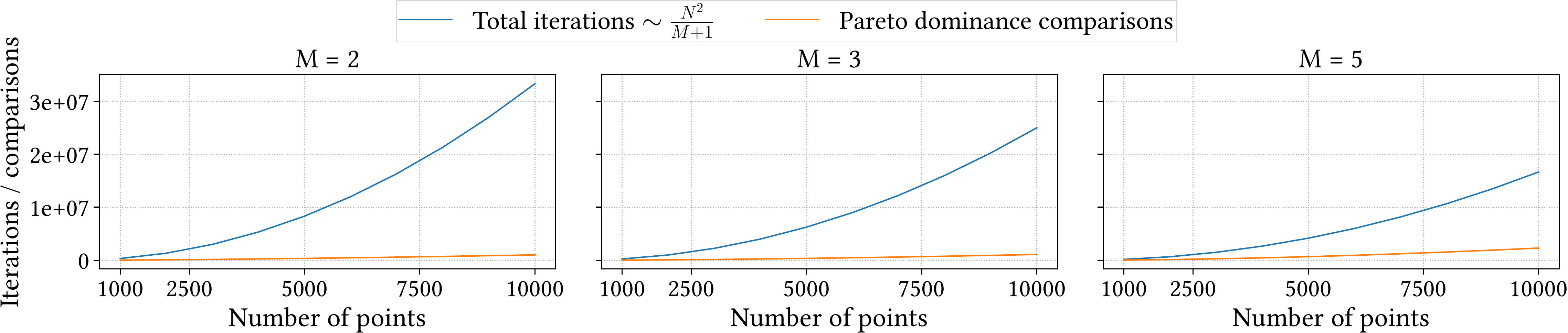}
\end{figure*}

    Figure~\ref{fig:rankord-complexity}, determined empirically by taking 1000-run averages of inner loop iterations and dominance checks of \emph{RankOrdinal}, shows that the number of dominance comparisons grows much slower than the total number of iterations.

\subsubsection{Example}\label{subsubsec:rankord-example}

We consider the case $N=10,~M=2$ with the following points:
\begin{align*}
    S_1 &= \begin{bmatrix} 0.79 & 0.35 \end{bmatrix} &
    S_2 &= \begin{bmatrix} 0.40 & 0.71 \end{bmatrix} &
    S_3 &= \begin{bmatrix} 0.15 & 0.014 \end{bmatrix}\\[-0.6ex]
    S_4 &= \begin{bmatrix} 0.46 & 0.82 \end{bmatrix} &
    S_5 &= \begin{bmatrix} 0.28 & 0.98 \end{bmatrix} &
    S_6 &= \begin{bmatrix} 0.31 & 0.74 \end{bmatrix} \\[-0.6ex]
    S_7 &= \begin{bmatrix} 0.82 & 0.52 \end{bmatrix} &
    S_8 &= \begin{bmatrix} 0.84 & 0.19 \end{bmatrix} &
    S_9 &= \begin{bmatrix} 0.85 & 0.78 \end{bmatrix} \\[-0.6ex]
    S_{10} &= \begin{bmatrix} 0.96 & 0.83 \end{bmatrix}
\end{align*}
The following permutations are obtained by stable-sorting according to the two objectives:
\begin{align*}
    \mathbf{p}_1 &= \begin{bmatrix} 3 & 5 & 6 & 2 & 4 & 1 & 7 & 8 & 9 & 10 \end{bmatrix}^T\\[-0.6ex]
    \mathbf{p}_2 &= \begin{bmatrix} 3 & 8 & 1 & 7 & 2 & 6 & 9 & 4 & 10 & 5 \end{bmatrix}^T
\end{align*}
The corresponding matrices $\mathbf{P}$ and $\mathbf{R}$ are:
\begin{align*}
    \mathbf{P} &= \begin{bmatrix}
        3 & 5 & 6 & 2 & 4 & 1 & 7 & 8 & 9 & 10\\
        3 & 8 & 1 & 7 & 2 & 6 & 9 & 4 & 10 & 5
    \end{bmatrix}^T\\
    \mathbf{R}& = \begin{bmatrix}
        6 & 4 & 1 & 5 & 2 & 3 & 7 & 8 & 9 & 10\\
        3 & 5 & 1 & 8 & 10 & 6 & 4 & 2 & 7 & 9
    \end{bmatrix}^T
\end{align*}

The algorithm will iterate over $\mathbf{p}_0$ and assert dominance between
each solution $S_i$ (highlighted with a circle at each iteration) and its
successors.
The successors $S_j$ that are examined in the inner loop are
highlighted with an underline. The conditions which apply at each inner loop
iteration are encoded as follows:
\begin{description}
    \item[$S_j^\star$] $\to$ $S_i \preceq S_j$ determined after a full rank comparison.
    \item[$S_j^\bullet$] $\to$ $S_i \npreceq S_j$ determined after a full rank comparison.
    \item[$S_j^\square$] $\to$ $S_j$ skipped due to $\text{rank}(i) \neq \text{rank}(j)$ (Alg.~\ref{alg:rankord} line 10)
\end{description}
The progress of the algorithm is illustrated in
Figure~\ref{fig:rankord-example}. The resulting fronts are:
$F_1 = \{ 3 \}, F_2 = \{ 1,2,5,6,8 \}, F_3 = \{ 4,7 \}, F_4 = \{ 9 \}, F_5 = \{ 10 \}$.

\begin{figure*}
    \caption{Step-by-step operation of the RankOrdinal and RankIntersect algorithms}\label{fig:examples}
    \begin{subfigure}{\textwidth}
        \centering
        \begin{tabular}{r|cccccccccc|cccccccccc}
    \# & \multicolumn{10}{|c}{Outer iteration $S_i$ (circled), inner iteration $S_j$ (underlined)} & \multicolumn{10}{|l}{Domination ranks} \\ 
    \hline
    1 & \cir{3} & 5 & 6 & 2 & 4 & 1 & 7 & 8 & 9 & 10 & 2 & 2 & 1 & 2 & 2 & 2 & 2 & 2 & 2 & 2\\
      & \cir{3} & \underline{8}$^\star$ & \underline{1}$^\star$ & \underline{7}$^\star$ & \underline{2}$^\star$ & \underline{6}$^\star$ & \underline{9}$^\star$ & \underline{4}$^\star$ & \underline{10}$^\star$ & \underline{5}$^\star$ & & & & & & & & & &\\
    \hline
    2 & 3 & \cir{5} & 6 & 2 & 4 & 1 & 7 & 8 & 9 & 10 & 2 & 2 & 1 & 2 & 2 & 2 & 2 & 2 & 2 & 2\\
      & 3 & 8 & 1 & 7 & 2 & 6 & 9 & 4 & 10 & \cir{5} & & & & & & & & & &\\
    \hline
    3 & 3 & 5 & \cir{6} & 2 & 4 & 1 & 7 & 8 & 9 & 10 & 2 & 2 & 1 & 3 & 2 & 2 & 2 & 2 & 3 & 3\\
      & 3 & 8 & 1 & 7 & 2 & \cir{6} & \underline{9}$^\star$ & \underline{4}$^\star$ & \underline{10}$^\star$ & \underline{5}$^\bullet$ & & & & & & & & & &\\
    \hline
    4 & 3 & 5 & 6 & \cir{2} & 4 & 1 & 7 & 8 & 9 & 10 & 2 & 2 & 1 & 3 & 2 & 2 & 2 & 2 & 3 & 3\\
      & 3 & 8 & 1 & 7 & \cir{2} & \underline{6}$^\bullet$ & \underline{9}$^\square$ & \underline{4}$^\square$ & \underline{10}$^\square$ & \underline{5}$^\bullet$ & & & & & & & & & &\\
    \hline
    5 & 3 & 5 & 6 & 2 & \cir{4} & 1 & 7 & 8 & 9 & 10 & 2 & 2 & 1 & 3 & 2 & 2 & 2 & 2 & 3 & 4\\
      & 3 & 8 & 1 & 7 & 2 & 6 & 9 & \cir{4} & \underline{10}$^\star$ & \underline{5}$^\square $ & & & & & & & & & &\\
    \hline
    6 & 3 & 5 & 6 & 2 & 4 & \cir{1} & \underline{7}$^\star$ & \underline{8}$^\bullet$ & \underline{9}$^\square$ & \underline{10}$^\square$ & 2 & 2 & 1 & 3 & 2 & 2 & 3 & 2 & 3 & 4\\
      & 3 & 8 & \cir{1} & 7 & 2 & 6 & 9 & 4 & 10 & 5 & & & & & & & & & &\\
    \hline
    7 & 3 & 5 & 6 & 2 & 4 & 1 & \cir{7} & \underline{8}$^\square$ & \underline{9}$^\star$ & \underline{10}$^\square$ & 2 & 2 & 1 & 3 & 2 & 2 & 3 & 2 & 4 & 4\\
      & 3 & 8 & 1 & \cir{7} & 2 & 6 & 9 & 4 & 10 & 5 & & & & & & & & & &\\
    \hline
    8 & 3 & 5 & 6 & 2 & 4 & 1 & 7 & \cir{8} & \underline{9}$^\square$ & \underline{10}$^\square$ & 2 & 2 & 1 & 3 & 2 & 2 & 3 & 2 & 4 & 4\\
      & 3 & 8 & 1 & 7 & 2 & 6 & 9 & 4 & 10 & 5 & & & & & & & & & &\\
    \hline
    9 & 3 & 5 & 6 & 2 & 4 & 1 & 7 & 8 & \cir{9} & \underline{10}$^\star$ & 2 & 2 & 1 & 3 & 2 & 2 & 3 & 2 & 4 & 5\\
      & 3 & 8 & 1 & 7 & 2 & 6 & 9 & 4 & 10 & 5 & & & & & & & & & &\\
\end{tabular}

        \caption{RankOrdinal Example}\label{fig:rankord-example}
    \end{subfigure}
    \begin{subfigure}{\textwidth}
        \centering
        \setlength{\tabcolsep}{3pt}
\begin{tabular}{l|llll|cccccccccc}
    \# & Perm. index & Rank & Dominance set & Rank set & \multicolumn{10}{l}{Domination ranks}\\
    \hline
    1 & $p_2(1) = 3$ & rank$(S_3)$ = 1 & $D(S_3) = \{1,2,4,5,6,7,8,9,10\}$ & $r_1 = \{ 3 \}$ & 2 & 2 & 1 & 2 & 2 & 2 & 2 & 2 & 2 & 2\\
    2 & $p_2(2) = 8$ & rank$(S_8)$ = 2 & $D(S_8) = \{9, 10\}$ & $r_2 = \{ 1,2,4,5,6,7,9,10 \}$ & 2 & 2 & 1 & 2 & 2 & 2 & 2 & 2 & 3 & 3\\
    3 & $p_2(3) = 1$ & rank$(S_1)$ = 2 & $D(S_1) = \{ 7 \}$ & $r_2 = \{ 1, 2, 4, 5, 6, 8 \}$ & 2 & 2 & 1 & 2 & 2 & 2 & 3 & 2 & 3 & 3\\
    4 & $p_2(4) = 7$ & rank$(S_7)$ = 3 & $D(S_7) = \{ 9, 10 \}$ & $r_3 = \{ 7 \}$ & 2 & 2 & 1 & 2 & 2 & 2 & 3 & 2 & 4 & 4\\
    5 & $p_2(5) = 2$ & rank$(S_2)$ = 2 & $D(S_2) = \{ 4 \}$ & $r_2 = \{ 1,2,5,6,8 \}$ & 2 & 2 & 1 & 3 & 2 & 2 & 3 & 2 & 4 & 4\\
    6 & $p_2(6) = 6$ & rank$(S_6)$ = 2 & $D(S_6) = \emptyset$ & $r_2 = \{ 1,2,5,6,8 \}$ & 2 & 2 & 1 & 3 & 2 & 2 & 3 & 2 & 4 & 4\\
    7 & $p_2(7) = 9$ & rank$(S_9)$ = 4 & $D(S_9) = \{ 10 \}$ & $r_4 = \{ 9 \}$ & 2 & 2 & 1 & 3 & 2 & 2 & 3 & 2 & 4 & 5\\
    8 & $p_2(8) = 4$ & rank$(S_4)$ = 3 & $D(S_4) = \emptyset$ & $r_3 = \{ 4, 7 \}$ & 2 & 2 & 1 & 3 & 2 & 2 & 3 & 2 & 4 & 5\\
    9 & $p_2(9) = 10$ & rank$(S_{10})$ = 5 & $D(S_{10}) = \emptyset$ & $r_5 = \{ 10 \} $ & 2 & 2 & 1 & 3 & 2 & 2 & 3 & 2 & 4 & 5\\
    10 & $p_2(10) = 5$ & rank$(S_5)$ = 2 & $D(S_5) = \emptyset$ & $r_2 = \{ 1, 2, 5, 6, 8 \} $ & 2 & 2 & 1 & 3 & 2 & 2 & 3 & 2 & 4 & 5\\
\end{tabular}

        \caption{RankIntersect Example}\label{fig:rankint-example}
    \end{subfigure}
\end{figure*}

\subsection{RankIntersect Algorithm}

The \emph{RankIntersect} (RS) algorithm takes a solution's dominance set as the
intersection of its objective-wise dominance sets:
\begin{align}
    D(S_i) = \bigcap_k D_k(S_i)
\end{align}

The intersections are efficiently computed by exploiting bit-level parallelism.
In a bitset, integer values are encoded as positions of the set bits. This has
the advantage of reducing set intersections to simple logical $\land$
operations that are optimized in the hardware.

The bitset approach is also employed by the MNDS~\cite{Moreno2020}, however
MNDS considers a dominance set as the set of solutions (or equivalently, their
permutation indices) that dominate the solution under consideration and updates
the considered solution's rank using the maximum rank in the dominance set (the
same rank update mechanism is also shared by best order sort~\cite{Roy2016}).
This leads to inefficiencies due to the overlap in dominance sets and the need
to compute maximum ranks.

According to Equation~\ref{eq:k-dominance-set}, for each solution $S_i$, \emph{RankIntersect} updates the ranks of successors $S_j$ where the current rank of $S_j$ equals the rank of $S_i$ and $S_j \succeq S_i$.


To implement this idea, \emph{RankIntersect} maintains a separate collection of
sets (bitsets) corresponding to each dominance rank value (from 1 to ...), such
that each set contains the permutation indices of individuals with that rank.
This rank set is updated along with the rank updates and used to reduce each
individual's dominance set according to Equation~\ref{eq:k-dominance-set}. The
resulting algorithm is illustrated in Algorithm~\ref{alg:rankint}:
\begin{itemize}

\item The first part of the algorithm up to line 11 is responsible for preprocessing
and initialization of data structures. A work bitset $b$ is used to initialize dominance sets $B(i)$ for each individual $i$.

\item Line 12 marks the beginning of the algorithmic loop where objective-wise dominance sets are intersected. When the last objective is reached, the algorithm proceeds to rank the rank assignment phase.

\item Lines 19--20 perform the bitset intersections including the additional intersection with the corresponding rank set.

\item The dominance set consisting of equally ranked successors is obtained at line 21. These successors will have their rank incremented at line 25. Therefore, they are removed from the current rank set (line 22) and added to the next one (line 23).

\end{itemize}

The resulting algorithm is functionally equivalent
with~\emph{RankOrdinal}~whose correctness has already been discussed. Relying
on bitset operations requires extra storage space for the bitsets but offers a
speed advantage as bit-level parallelism can accelerate intersection operations
with a factor equal to the basic data block size (i.e., a 64-bit integer used
as a basic data block can store 64 permutation indices), which can be further
increased by vectorization.

\begin{algorithm}
    \small
    \caption{\emph{RankIntersect}}\label{alg:rankint}
        \SetKwFunction{rankInt}{RankIntersect}
    \SetKwFunction{stableSort}{StableSort}
    \KwIn{$S_1,...,S_N$}
    \KwOut{Pareto fronts $F_1, F_2, ... $}
    \SetKwProg{Fn}{Function}{:}{}
    \Fn{\rankInt{$S_1,...,S_N$}}{
        $\mathbf{P} \gets$~\stableSort{$S_1,...,S_N$}\; 
        $\mathbf{B} \gets$ list of bitsets associated to each $S_i$\;
        $\mathbf{K} \gets$ list of bitsets associated to each dominance rank $r$\;
        $\mathbf{K}(1) \gets \text{new bitset}$\;
        set all bits in $\mathbf{K}(1)$\tcp*{all solutions start with rank 1}
        $b \gets $ auxiliary bitset used to compute the set intersections\;
        set all bits in $b$\;
        \ForEach{i \upshape{from} 1 \upshape{to} N}{
            reset bit $i$ in $b$\;
            $B(i) \gets b$\;
        }

        \ForEach{\upshape{$k$ from $2$ to $M$}}{
            set all bits in $b$\;
            \ForEach{\upshape{$i$} in $\mathbf{p}_k$}{
                reset bit $i$ in $b$\;

                \uIf{$k < M$}{
                    $\mathbf{B}(i) \gets \mathbf{B}(i) \land b$\;
                }
                \Else{
                    $r \gets \mathbf{K}\left(\text{rank}(i)\right)$\tcp*{bitset for current rank}
                    $s \gets \mathbf{K}\left(\text{rank}(i+1)\right)$\tcp*{bitset for next rank}
                    $v \gets \mathbf{B}(i) \land b \land r$\;
                    $r \gets r \land \lnot v$\tcp*{remove dominance set from $r$}
                    $s \gets s \lor v$\tcp*{add dominance set to $s$}
                    
                    \ForEach{\upshape{$j$} in $v$}{
                        $\text{rank}(j) = \text{rank}(j)+1$\;
                    }
                }
            }
        }

        $\mathbf{F} \gets$ list of fronts of size $\max_i\text{rank}(i$)+1\;
        \ForEach{\upshape{$i$ from $1$ to $N$}}{
            append $i$ to front $F_{\text{rank}(i)}$\;
        }
        \textbf{return} $\mathbf{F}$\;
    }

\end{algorithm}

\subsubsection{Computational complexity}

Since \emph{RankIntersect} does not perform any dominance checks we discuss computational complexity in terms of the costs of computing the necessary set intersections.

\subsubsection*{Worst case}
Logical operations $\land, \lor$ over bitsets are performed by iterating over $\lceil \frac{N}{64} \rceil $ basic data blocks which are internally stored in a list by the bitset. This is due to the fact that a data block is represented by a 64-bit unsigned integer. The outer loop at line 14 performs $N$ iterations, while the inner loop performs $\lceil \frac{N}{64} \rceil$ iterations to intersect the bitsets. Therefore the total number of iterations will be in the order of $N^2$ which leads to $O(M N^2)$ worst-case complexity.

\subsubsection*{Best case}
Since Algorithm~\ref{alg:rankint} successively intersects bitsets going from one objective to the next (lines 12, 17), it becomes possible to memorize the regions at the beginning and end of the list where the data blocks have become zero after the intersection and then skip them in the next iteration. This creates a short-circuiting mechanism for early stopping in the case of non-dominance. When all solutions are non-dominated, as all the dominance sets will be empty, the remaining complexity is $O(M N \log N)$ given by the sorting phase.

\smallskip
We conclude that the overall complexity of \emph{RankIntersect} is $O(M N^2)$ in the worst case and $O(M N\log N)$ in the best case. However, the advantage of \emph{RankSort} comes from the fact that its operations are very basic and can exploit the characteristics of modern processors. For $M > 2$, most runtime effort is spent performing objective-wise dominance set intersections at line 17. However in practice, a relatively modern CPU can intersect four basic data blocks at a time using AVX2 instructions (since a ``wide'' SIMD type can hold four 64-bit unsigned integers). For large $N$, the runtime effort shifts towards the sorting phase, suggesting that additional runtime benefits might be achieved by parallelizing the sorting.

%

\subsubsection{Example}
We consider the same input from Section~\ref{subsubsec:rankord-example}
consisting of $N=10$ points with $M=2$. The matrix $\mathbf{P}$ was:
$$
    \mathbf{P} = \begin{bmatrix}
        3 & 5 & 6 & 2 & 4 & 1 & 7 & 8 & 9 & 10\\
        3 & 8 & 1 & 7 & 2 & 6 & 9 & 4 & 10 & 5
    \end{bmatrix}^T
$$
The algorithm will iterate over $\mathbf{p}_2$ (the last column of
$\mathbf{P}$) in the order: 3, 8, 1, 7, 2, 6, 9, 4, 10, 5. Since initially all
individuals start at rank one, the first rank set $r_1$ will contain all the
permutation indices. Each subsequent rank set is initialized with $\emptyset$.
Then, the dominance set and rank sets are updated as shown in
Figure~\ref{fig:rankint-example}. As their ranks get updated, the permutation
indices of the respective solutions are removed from the current rank set and
inserted into the next one. Note that the dominance sets and rank sets shown in
Figure~\ref{fig:rankint-example} represent the final values resulting after the
assignments at lines 19 and 20, respectively, in Algorithm~\ref{alg:rankint}.


\section{Benchmark Results}\label{sec:results}

The effectiveness of the proposed algorithms is measured both synthetically and
in a practical optimization setting with the help of the \emph{Pagmo}
framework~\citep{Biscani2020}, a C\verb!++! library for massively parallel
optimization that provides a unified interface to optimization algorithms and
problems.

We extended \emph{Pagmo} with C\verb!++! implementations for the tested
algorithms and ran experiments with the DTLZ benchmark set~\citep{Deb2005}
(problems DTLZ1 to DTLZ5) using the NSGA2 algorithm~\citep{Deb2002}.

All tests were performed on a workstation containing an AMD
Ryzen\texttrademark{} 5950X CPU with 8Mb of L2 cache and 64Mb of L3 cache, and
64Gb of DDR4 3600Mhz-CL16 memory. The experiments were run on a single thread
to minimize effects such as resource contention or CPU pipeline stalls and to
be able to acquire accurate profiling results.
The code was compiled on GNU/Linux using the GNU C\verb!++! compiler version
11.2.0 with compilation flags: \texttt{-O3 -mavx2 -mfma}.

Each tested algorithm was implemented according to descriptions and pseudocode
made available in their respective publications. A reasonable amount of effort
has been expended for profiling and tuning each implementation. In the case of
MNDS, the code has been ported from the original Java implementation in jMetal.

Since not all algorithms handle duplicate solutions in the same way, the
procedure for handling duplicates in the population was abstracted away and
made common between implementations. This ensures consistency and
reproducibility among algorithms such that, regardless of the concrete
non-dominated sorting implementation, the same random seed leads to the same
results.

\begin{figure*}[ht]
    \caption{Pagmo DTLZ1 -- elapsed time (seconds)
    averaged over five algorithmic runs}\label{fig:pagmo-all}
    \includegraphics[width=0.9\textwidth]{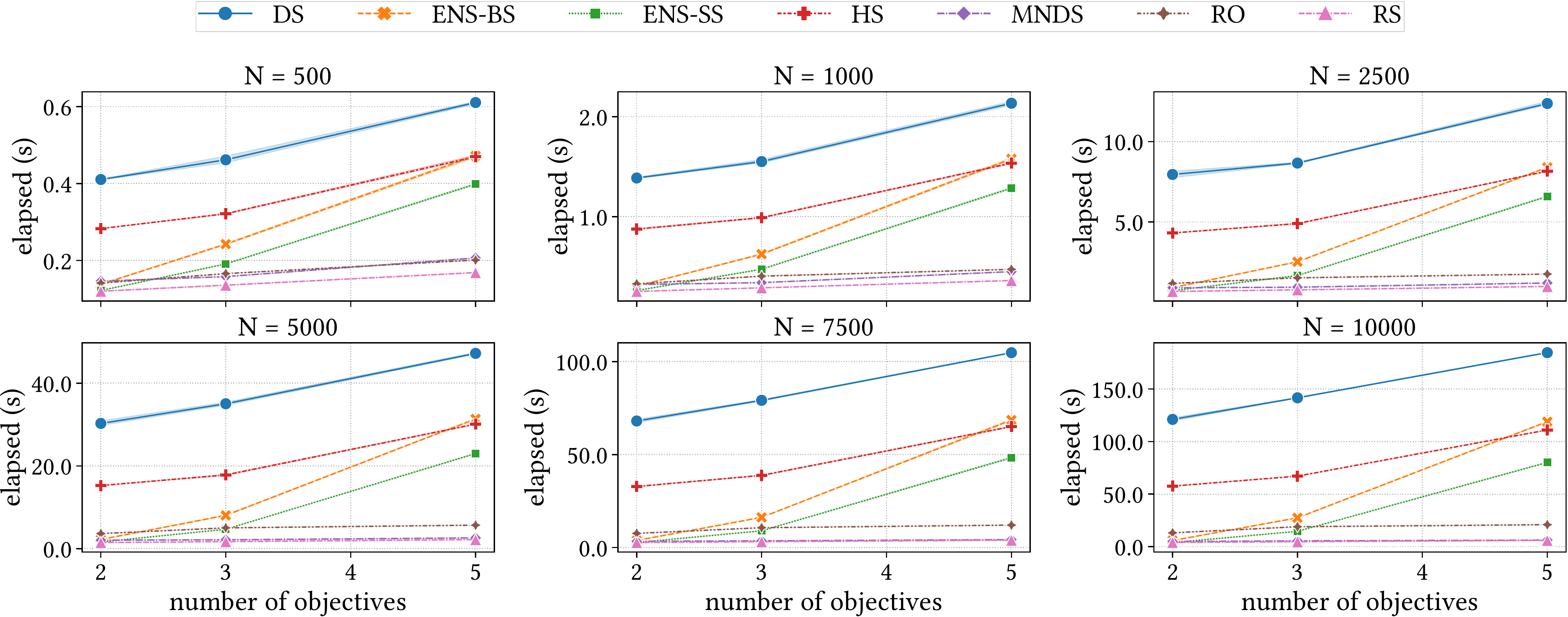}
\end{figure*}

Due to runtime and space constraints, we first compare a larger selection of
algorithms, consisting of RO, RS, MNDS, ENS-BS, ENS-SS, DS and HS on the DTLZ1
test problem, using 5 repetitions of the NGSA2 algorithm. We report the average
elapsed time over the 5 algorithmic runs (Figure \ref{fig:pagmo-all}).

We then focus on a direct comparison between the three fastest algorithms, RO, RS and MNDS,
where we report the average elapsed time over all runs (20 repetitions) and all problems (DTLZ1 to DTLZ5).

The NSGA2 algorithm was parameterized as follows:
\begin{itemize}
    \item Population size: 500, 1000, 2500, 5000, 7500, 10000
    \item Maximum generations: 100
    \item Problem dimension: 25
    \item Crossover rate: 95\%
    \item Crossover distribution rate $\eta_c$: 10
    \item Mutation rate: 10\%
    \item Mutation distribution rate $\eta_m$: 50
\end{itemize}

The results indicate that bitset-based approaches RS and MNDS are the overall
fastest, followed by RO. In the case of two objectives, the Efficient
Non-dominated Sort family of algorithms (ENS-SS and ENS-BS) also performs very
well, being similarly fast to RS and MNDS and marginally ahead of RO. This is a
remarkable result especially considering their simplicity. However, as the
number of objectives increases ($M=3$, $M=5$) these two algorithms become
slower due to the overhead of direct dominance comparisons within their
respective search strategy. The elapsed time for each algorithm with varying
number of objectives is shown in Figure~\ref{fig:pagmo-all}.



We perform a detailed comparison of the fastest algorithms, RO, RS and MNDS, on
the DTLZ benchmark set consisting of problems DTLZ1 to DTLZ5. Each algorithm
was tested with the same random seeds over 20 repetitions for each
configuration. The number of objectives was varied from 1 to 10 with an
increment of 1 and from 10 to 20 with an increment of 2.

\begin{figure*}[ht]
    \caption{RO, RS, MNDS -- Pagmo DTLZ benchmark set -- elapsed time (seconds) averaged over twenty runs and five problems (DTLZ1 to DTLZ5}\label{fig:pagmo-rs-mnds-elapsed}
    \includegraphics[width=0.9\textwidth]{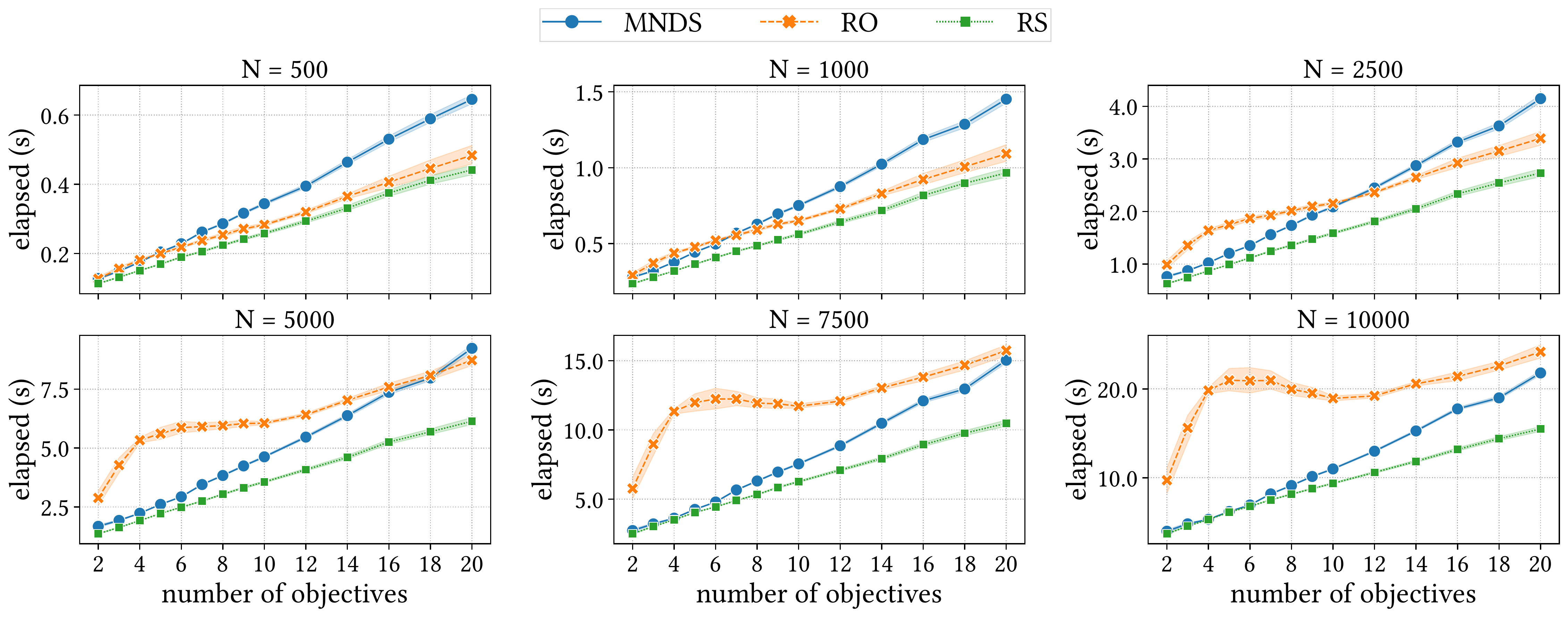}
\end{figure*}

Figure~\ref{fig:pagmo-rs-mnds-elapsed} shows that \emph{RankIntersect} provides
a consistent runtime benefit, finishing ahead of MNDS in all cases. \emph{RankOrdinal} performs better than MNDS for small populations but becomes slower as the population size increases ($N > 5000$), exhibiting an unexpected performance decrease for $M < 10$. This aspect is not yet fully understood, but we suspect some inefficiency at the level of the ordinal rank comparison, also considering the fact that the size of the dominance set is inversely proportional with $M$.

\begin{figure*}[ht]
    \caption{RO, RS, MNDS -- Synthetic benchmark randomly uniform objective values}\label{fig:synthetic-elapsed}
    \includegraphics[width=0.9\textwidth]{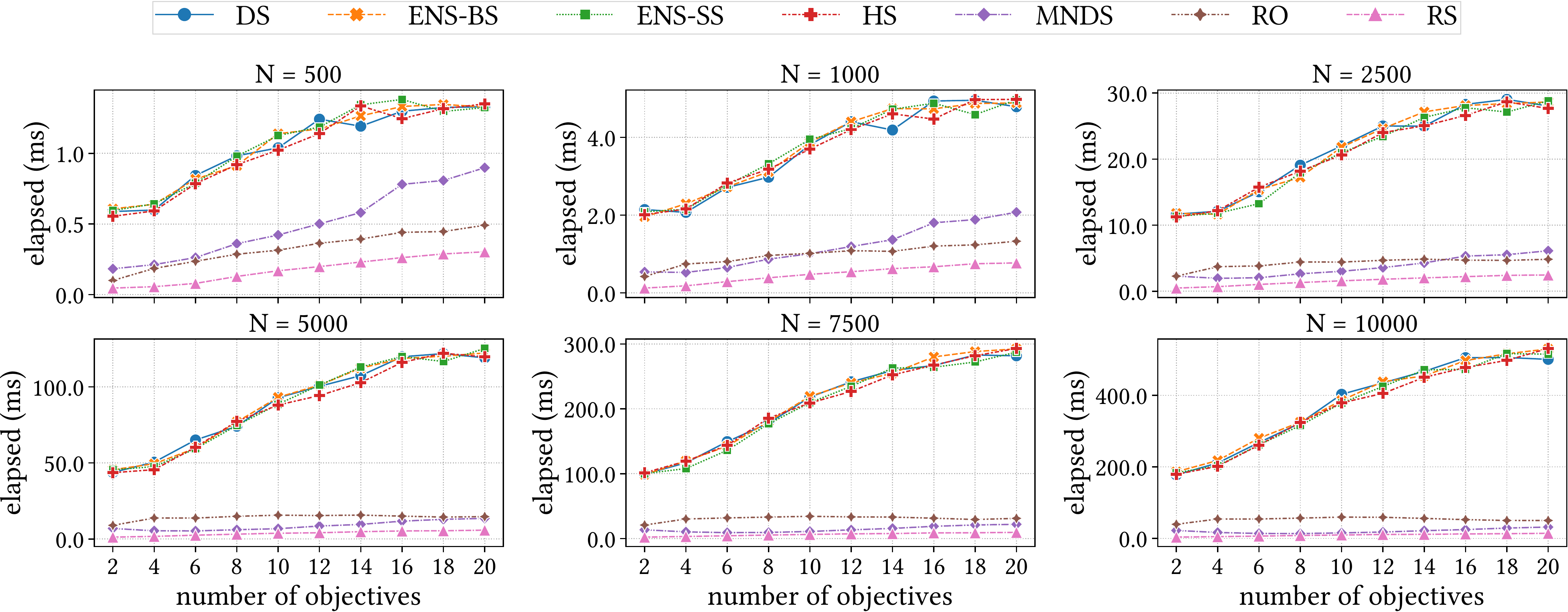}
\end{figure*}

Next, we tested the performance of all algorithms on a synthetic benchmark where the objective values are uniformly sampled from the unit hypercube. The results shown in Figure~\ref{fig:synthetic-elapsed} again show that RO is slightly faster than MNDS for lower population sizes but becomes slower as population size increases ($N > 5000$). The same slowdown for $M<10$ is observed for RO on synthetic data. The other algorithms are noticeably slower  than the trio RO, RS, MNDS but perform similarly well to each other. As the points are random, their respective ranking strategies are dominated by the runtime cost of Pareto dominance checks.

\section{Conclusion}\label{sec:conclusion}

In this paper we introduced two simple and performant algorithms for
non-dominated sorting, one of the most runtime-intensive components of
Pareto-based MOEAs.

In contrast to similar approaches BOS or MNDS, Rank Sort defines a dominance set as
the set of solutions dominated by the current solution (and not dominating the
current solution). The main insight leading to the better performance of
the Rank Sort algorithms is that a solution's rank needs only be updated when the
solution is found to be dominated by another solution of the same rank. This
further reduces the size of the dominance set that needs to be examined.

The first algorithm, \emph{RankOrdinal}, uses the comparison of ordinal ranks
to establish dominance. Instead of performing set intersection operations to
compute dominance sets, it simply iterates over the smallest objective-wise
dominance set. \emph{RankOrdinal} is slower than \emph{RankIntersect} and MNDS
but requires only $O(N)$ space. It is also slightly slower than ENS-SS and
ENS-BS in the two-objective case, but outperforms them as the number of
objectives increases.

The second algorithm, \emph{RankIntersect}, uses set intersections to compute
dominance sets. These are efficiently implemented using bit-level parallelism,
which is the main factor in its performance.

The two algorithms have $O(MN\log N)$ best-case and $O(MN^2)$ worst-case asymptotic complexity, in line with other non-dominated sorting algorithms. As already apparent from
Algorithms~\ref{alg:rankord} and~\ref{alg:rankint}, the proposed methods are
simple and easy to include in other frameworks. They use only basic data structures and have low constant factors.
Nevertheless, C\verb!++! implementations are provided for all the methods
tested in this paper.

Future work will focus on a better understanding of the computational
complexity of the proposed methods and on exploring other algorithmic ideas.
For example, performance can likely be improved by hybridizing the two steps of
the algorithm: stable sort and rank update into a single hybrid step in order
to further minimize the number of operations.

It will also be interesting to profile each algorithm and express its complexity in terms of low-level hardware events (executed instructions, branches, branch misses, cache misses). This will serve to provide a unified framework for the analysis of asymptotic complexity.

We also plan to provide efficient implementations of other state of the art non-dominated sorting algorithms and to publish all the algorithms as a stand-alone open-source library.

\bibliographystyle{plainnat}
\bibliography{references}

\appendix

\end{document}